# FUZZY SELECTIVE STATE SPACE MODEL FOR FORECASTING THE DYNAMICS OF CONSUMER PREFERENCES

**P.A. Golovinski[1], A.A. Drobyshev[2], A.O. Shatalova[3]**

[1]Laboratory of Cyber-Physical Systems, Voronezh State University
[2]Voronezh State University
[3]Voronezh State Technical University

**Abstract.** We consider forecasting the dynamics of consumer preferences of retail chain customers. A customer's preference is described by a triangular fuzzy number. The evolution of the preference over time is governed by a Mamba-type selective state space model whose hidden state is a triangular number, with computations carried out in M.G. Matveev's two-component algebra of fuzzy numbers. The recurrence parameters are computed from crisp features, while the fuzzy part of the model remains linear with non-negative coefficients. We show that the state then always remains a proper triangular number and is a convex combination of past observations, so that the fuzziness of the forecast does not accumulate over time and the algebraic computations coincide with Zadeh's extension principle. On a synthetic example with competitor promotions and customer switching, a lightweight 107-parameter model reaches a mean absolute error of the forecast centre of 0.154 versus 0.163 for the constant-inertia evolution operator and 0.179 for the naive forecast, matches a crisp model of the same architecture and additionally provides a linguistically interpretable estimate of forecast vagueness. We also show how the linguistic readout of the forecast can be used to detect a permanent switch of a customer to a competitor.

## 1. Introduction

Through loyalty programmes, retail chains accumulate the purchase histories of millions of customers. Customer preferences for product categories, brands and price segments do not stay constant. They change under the influence of prices and promotions, seasonality, the appearance of new products and changes in the customer's life circumstances. Forecasts of these changes underlie personalised offers, assortment planning and customer retention, so the quality of managerial decisions depends on how accurate and understandable such forecasts are.

The difficulty of the problem has two sides. The first concerns the nature of the notion of preference itself. A customer does not know his or her loyalty share for a brand to within a percent. The customer will say that he or she usually buys this brand but sometimes tries others. A marketer reasons in the same terms and divides customers into loyal, hesitant and leaving ones. L. Zadeh introduced the notion of a linguistic variable precisely to describe judgements of this kind, whose values are words rather than numbers [1], and later formulated the programme of computing with words [2]. According to fuzzy-trace theory [3], in decision making people rely mainly on the qualitative gist of information rather than on its verbatim quantitative details. H. Simon's concept of bounded rationality [4] leads to the same conclusion from another direction, showing that real choices are based on simplified approximate assessments. Therefore a point forecast of a preference hides essential information, while a probabilistic forecast describes uncertainty as random scatter, although to a large extent it stems from the vagueness of the concept and from the lack of information about the customer. A model that stores and outputs the preference in an approximate linguistic form is convenient for the decision maker [5].

The simplest form of representing an approximate estimate is the triangular fuzzy number $\tilde{x} = (l, m, r)$, read as "about $m$, but not less than $l$ and not more than $r$". The same three-point estimate, comprising the optimistic, most likely and pessimistic values, has long been used in expert planning, for example in the PERT method [6]. The parameters of a triangular number have a direct meaning. The mode $m$ gives the typical level of preference, and the width $r - l$ reflects how vague our knowledge of the customer is; it grows when observations are scarce or behaviour is inconsistent.

The second side of the problem is computational. The forecast is built from the customer's purchase history, which is a long, irregular and continuously growing sequence of events. In recommender systems, such sequences were first processed with recurrent neural networks [7], which were then largely superseded by transformers [8], for example the SASRec model [9]. The core of a transformer is the attention mechanism, in which every element of the sequence is compared with all previous ones. Hence the amount of computation grows quadratically with the length of the history, and the whole customer history has to be stored to make a forecast. For a retail chain with millions of customers whose histories grow every week, this is a serious limitation. There is also a second reason, important for our problem. Elements in the attention mechanism are compared through scalar products followed by exponential normalisation, that is, by means of the softmax function. If the sequence elements are fuzzy, such operations take the result out of the class of triangular numbers, and the fuzziness would have to be approximately restored at every layer of the model. Selective state space models, above all Mamba [10, 11], are free of both drawbacks. They are trained in time linear in the sequence length, update the state in constant time during step-by-step inference, and, as will be seen below, their recurrent core is linear in the state. These models are already used in sequential recommender systems [12, 13] and in time series forecasting [14, 15]; however, they produce crisp forecasts, and the known ways of estimating uncertainty in them are probabilistic [15].

For the problem of the dynamics of consumer preferences, a state space model is also natural in substance. A customer does not change preferences abruptly at every purchase. The current preference is formed by past experience that is gradually forgotten, which is exactly described by the linear recurrence

$$h_t = \bar{A}_t h_{t-1} + \bar{B}_t x_t, \qquad y_t = C_t h_t, \tag{1}$$

in which $h_t$ is the hidden state, $x_t$ is the observation, and the coefficients $\bar{A}_t$ and $\bar{B}_t$ set the inertia and the susceptibility to new experience. The simplest preference evolution operator with constant inertia, $h_t = \lambda h_{t-1} + (1-\lambda)x_t$, is a special case of (1). In Mamba the coefficients depend on the input through the discretisation step $\Delta_t$, so that $\bar{A}_t = \exp(\Delta_t A)$ with $A < 0$. For preferences this means an inertia that depends on circumstances. Before a competitor's promotion or in a season, the model can forget the past faster, which an operator with constant inertia cannot do.

The main idea of the paper is to combine a fuzzy description of preferences with a selective state space model. The recurrent core (1) is linear in the state, and linear operations with non-negative crisp coefficients map triangular numbers into triangular numbers. If all recurrence coefficients are computed from crisp features and their sign is controlled, the fuzzy state propagates in time without any approximation. A suitable mathematical apparatus for such computations is the two-component algebra of fuzzy numbers developed by M.G. Matveev and his students [16, 17, 18, 19, 20]. Section 2 describes this algebra and identifies the operations that preserve the class of triangular numbers. Section 3 constructs the fuzzy selective model, which we call FW-Mamba (Fuzzy W-algebra Mamba), studies its properties and describes the training procedure. Section 4 contains a computational experiment on synthetic data, including the detection of a permanent customer switch, and Section 5 summarises the results and outlines further work.

## 2. Triangular fuzzy numbers and the two-component algebra

We describe a preference by a triangular fuzzy number $\tilde{x} = (l, m, r)$ whose membership function increases linearly from zero at $l$ to one at $m$ and then decreases linearly to zero at $r$. Earlier we used such numbers to describe fuzzy user ratings in a recommender system for choosing a learning trajectory [21], where arithmetic operations on triangular numbers and the metric used below as the loss function were introduced. A number is called proper if $l \le m \le r$, so that the membership function has the usual shape.

For further computations it is convenient to represent numbers in parametric form, that is, through $\alpha$-cuts, the sets of values whose membership degree is not less than $\alpha$. For a triangular number these are the intervals

$$\tilde{x}_\alpha = [x_L(\alpha), x_R(\alpha)], \qquad 0 \le \alpha \le 1, \tag{2}$$

whose endpoints depend linearly on $\alpha$:

$$x_L(\alpha) = l + (m - l)\,\alpha, \qquad x_R(\alpha) = r - (r - m)\,\alpha. \tag{3}$$

The function $x_L$, inverse to the increasing (left) branch of the membership function, is called the left component of the number, and the function $x_R$, inverse to the decreasing (right) branch, is called the right component. Each component is linear and is therefore uniquely determined by its two values at $\alpha = 0$ and $\alpha = 1$: the left component by the pair $(l, m)$ and the right one by the pair $(r, m)$. At $\alpha = 1$ the components coincide, $x_L(1) = x_R(1) = m$. Thus the triple $(l, m, r)$ is merely a shorthand for the pair of components $(x_L, x_R)$, and it is used below in exactly this sense.

Classical fuzzy arithmetic is based on Zadeh's extension principle [22] and reduces to interval arithmetic on $\alpha$-cuts. This arithmetic has well-known drawbacks that become essential in long chains of computations. In it, the difference of a number and itself is not zero, the fuzziness of the result grows with every operation, and the product of two triangular numbers is no longer a triangular number.

In the two-component algebra (W-algebra) [17, 18, 19] a number is regarded as a pair of components $(x_L, x_R)$, and operations are performed independently on the left components of the operands, with the result written to the left component, and on the right components, with the result written to the right one. For the result to be a triangular number again, the operation on components is defined so as to preserve their linearity in $\alpha$. An operation $* \in \{+, -, \cdot, /\}$ is performed on the values of the components at $\alpha = 0$ and $\alpha = 1$ according to the rules for real numbers, and between them the result is completed linearly [17, Def. 9]:

$$(x_1 * x_2)(\alpha) = x_1(0) * x_2(0) + [x_1(1) * x_2(1) - x_1(0) * x_2(0)]\,\alpha. \tag{4}$$

Since the values of the left component at $\alpha = 0$ and $\alpha = 1$ are $l$ and $m$, and those of the right component are $r$ and $m$, in triple notation operations (4) become componentwise [17, Def. 11]:

$$\begin{gathered}\tilde{x}_1 \oplus \tilde{x}_2 = (l_1 + l_2,\ m_1 + m_2,\ r_1 + r_2), \qquad \lambda \odot \tilde{x} = (\lambda l,\ \lambda m,\ \lambda r), \\ \tilde{x}_1 \odot \tilde{x}_2 = (l_1 l_2,\ m_1 m_2,\ r_1 r_2),\end{gathered} \tag{5}$$

where a crisp number $c$ is identified with $(c, c, c)$.

For addition and multiplication by a crisp number, rule (4) coincides with the usual pointwise addition of the functions $x_L$, $x_R$ and their multiplication by a number, since these operations preserve linearity by themselves. For the multiplication of two fuzzy numbers this is not so, and it is important not to confuse three different objects here. The pointwise product of components $x_{1L}(\alpha)\, x_{2L}(\alpha)$ is quadratic in $\alpha$. For non-negative numbers it coincides with the product by the extension principle, whose $\alpha$-cut endpoints are $x_{1L}x_{2L}$ and $x_{1R}x_{2R}$; its membership function has curved sides, that is, the result is no longer triangular. The W-product (5) coincides with this pointwise product at $\alpha = 0$ and $\alpha = 1$, that is, at the vertices $l_1 l_2$, $m_1 m_2$, $r_1 r_2$, and straightens the sides of the triangle. Thus the multiplication formula in (5) is neither the Zadeh product nor the pointwise product of the membership function branches. It is a definition of the operation in the algebra, chosen so that the set of triangular numbers is closed under multiplication and division is inverse to multiplication. In the model constructed below, fuzzy numbers are never multiplied by each other, so this difference does not affect the results.

Operations (5) turn the space of triples into a commutative ring in which every number has an opposite, and $\tilde{x} \ominus \tilde{x} = 0$. Computations in the algebra reduce to three parallel real-valued computations, which allows standard computational schemes to be carried over to fuzzy data.

This is how in [20] the problem of fuzzy linear regression with crisp parameters is reduced to the least squares method.

The price of this algebraic simplicity is that not all operations preserve properness. Multiplication by a negative number reverses the order of the components and produces a triple with $l > m > r$, which no longer describes an ordinary membership function. It is therefore important to understand which operations certainly do not lead out of the class of proper numbers. Let $\tilde{x}_1$ and $\tilde{x}_2$ be proper and $\lambda_1, \lambda_2 \geq 0$. Then the inequalities $l_i \leq m_i \leq r_i$ are preserved under multiplication by $\lambda_i$ and add up, so the combination $\lambda_1 \odot \tilde{x}_1 \oplus \lambda_2 \odot \tilde{x}_2$ is proper as well. In other words, proper numbers form a convex cone closed under addition and multiplication by non-negative numbers. The same holds for a shift by any crisp number, which changes neither the order of the components nor the width.

It is essential that for these operations the two-component algebra gives the same result as Zadeh's extension principle. Indeed, the $\alpha$-cut of a non-negative linear combination of non-interacting fuzzy numbers is the same combination of the intervals (2). With non-negative coefficients, left endpoints add with left ones and right endpoints with right ones, and the endpoints of the result remain linear functions of $\alpha$. This means that the result is again a triangular number and its vertices are obtained componentwise, as in (5). Thus, in the part of the model where only such operations are used, Matveev's algebra is not an approximation to classical fuzzy arithmetic but coincides with it. A discrepancy arises only with negative coefficients and with the multiplication of fuzzy numbers, and we construct the model so that the fuzzy computational path contains neither.

A similar argument applies to nonlinear functions. If a function $\varphi$ is non-decreasing, its componentwise application $\varphi(\tilde{x}) = \big(\varphi(l), \varphi(m), \varphi(r)\big)$ preserves the order of the components. By the extension principle, a non-decreasing continuous function maps the $\alpha$-cut $[u, v]$ onto the interval $[\varphi(u), \varphi(v)]$, so the values at the levels $\alpha = 0$ and $\alpha = 1$, i.e. the vertices of the triangle, are computed exactly. The only difference is the shape of the sides between the vertices, which may become curved under Zadeh's principle but stay straight in the algebra. The model built in this paper uses only one such function, namely clipping to the interval $[0,1]$, which is needed for the preference to remain a share.

## 3. Fuzzy selective state space model

Suppose that for a customer at every week $t$ we know a fuzzy observation $\tilde{x}_t$ obtained from the customer's purchases and a vector of crisp features $s_t$. The features include the centre $c(\tilde{x}_t) = (l_t + m_t + r_t)/3$, i.e. the abscissa of the centroid of the triangle under the membership function, and the width $w(\tilde{x}_t) = r_t - l_t$ of the current observation, as well as external factors. In the example considered in Section 4, these features are a competitor's promotion in the current week and in the next, pre-announced week, a seasonal component and the logarithm of the number of purchases. The task is to build from the observation history a fuzzy forecast $\tilde{p}_{t+1}$ of the next week's observation.

The key decision is to separate the roles of the crisp and fuzzy parts of the model. All nonlinearities that make the model selective are applied only to crisp features and produce non-negative crisp coefficients. The fuzzy path from observations to the forecast consists only of addition, multiplication by non-negative numbers, shift and clipping. In substance, the crisp part decides how much the representation of the customer should be updated, while the fuzzy part stores information about the customer together with a measure of its imprecision. According to Section 2, with this construction all fuzzy computations are exact.

Consider first a single hidden state. The discretisation step is computed from crisp features as $\Delta_t = \text{softplus}\left(w_\Delta^\top s_t + b_\Delta\right)$ and is therefore always positive. We assume that the continuous dynamics of the state is described by the equation $\dot{h} = a(h - x)$ with $a < 0$, that is, the state relaxes towards the current observation. The exact zero-order-hold discretisation of this equation with step $\Delta_t$, which is also used in Mamba, gives

$$\tilde{h}_t = \bar{a}_t \odot \tilde{h}_{t-1} \oplus (1 - \bar{a}_t) \odot \tilde{x}_t, \qquad \bar{a}_t = e^{a\Delta_t} \in (0,1). \tag{6}$$

This is the Mamba recurrence (1) with $\bar{B}_t = 1 - \bar{A}_t$, which corresponds to the choice $B = -A$ in the continuous model. Both coefficients in (6) are non-negative, so by induction the state remains a proper triangular number provided the initial state and all observations are proper.

Unrolling recurrence (6), we obtain an explicit expression of the state through the observation history

$$\tilde{h}_t = \omega_{t,0} \odot \tilde{h}_0 \oplus \bigoplus_{\tau=1}^{t} \omega_{t,\tau} \odot \tilde{x}_\tau, \qquad \omega_{t,\tau} = (1 - \bar{a}_\tau) \prod_{k=\tau+1}^{t} \bar{a}_k\,, \qquad \omega_{t,0} = \prod_{k=1}^{t} \bar{a}_k. \tag{7}$$

All weights $\omega_{t,\tau}$ are non-negative and sum to one, which is easily verified by induction since $\bar{a}_t \cdot 1 + (1 - \bar{a}_t) = 1$. Consequently, the state is a convex combination of past observations with weights that decay exponentially with age, the rate of decay being set by the circumstances of each week through $\Delta_t$. Two important corollaries follow immediately from (7). First, if all observations lie in $[0,1]$, so does the state. Second, the width of the state equals the same convex combination of observation widths,

$$w(\tilde{h}_t) = \sum_\tau \omega_{t,\tau}\, w(\tilde{x}_\tau) \le \max_\tau w(\tilde{x}_\tau), \tag{8}$$

and therefore cannot exceed the largest observation width. This property fundamentally distinguishes the model from fuzzy recurrent schemes with fuzzy coefficients, in which the width grows at every step and the forecast becomes uninformative after a few steps. Here uncertainty is remembered exactly as long as the observations themselves are remembered. Note also that each observation enters (7) only once, so the dependency effect that inflates widths in interval arithmetic does not arise.

A single state with a single memory time is not flexible enough; therefore, as in Mamba, the model contains $D$ channels with $N$ states each, with different constants $a_{d,n} < 0$ and channel-specific steps $\Delta_{t,d}$. All channels receive the same observation $\tilde{x}_t$ and differ in their memory times. Each state obeys recurrence (6), and the output is formed in two stages. First, within each channel the states are mixed with weights $C_t = \mathrm{softmax}(W_C s_t)$ that select the required memory time scale; then the channels are mixed with weights $\pi_t = \mathrm{softmax}(W_\pi s_t)$, and a crisp shift $\beta_t = w_\beta^{\mathrm{T}} s_t + b_\beta$ is added to the result. The forecast reads

$$\tilde{p}_{t+1} = \Pi_{[0,1]}\big(\oplus_{d=1}^{D} \pi_{t,d} \odot \oplus_{n=1}^{N} C_{t,n} \odot \tilde{h}_{t,d,n} \ \oplus \ \beta_t\big), \tag{9}$$

where $\Pi_{[0,1]}$ denotes componentwise clipping to $[0,1]$. The weights $C_t$ and $\pi_t$ are non-negative and sum to one, so the expression in brackets without the shift is again a convex combination of observations and inherits both properties obtained above. The shift $\beta_t$ describes the additive influence of external factors on the preference level and does not change the width. Clipping is non-decreasing and, according to Section 2, preserves properness. Thus forecast (9) is a proper triangular number in $[0,1]$ for any parameter values.

With $D = N = 4$ and six features, the model has 107 trainable parameters. This is a lightweight version of Mamba without high-dimensional input and output projections, the convolutional layer and the multilayer structure. The lightness here is not only computational. A small number of parameters with clear meaning allows the trained model to be interpreted directly, as will be seen in Section 4.

Training uses the fuzzy observations of the next step. As a closeness measure it is natural to take the distance in the two-component algebra, that is, the mean squared deviation over the three components

$$\mathcal{L} = \frac{1}{3|\mathcal{M}|} \sum_{(i,t)\in\mathcal{M}} \| \tilde{p}_{t+1}^{(i)} - \tilde{x}_{t+1}^{(i)} \|^2, \tag{10}$$

where the sum runs over the customer–week pairs for which the observation is informative, that is, the customer made purchases in the category. This is the squared distance between triangular numbers introduced in [21]. The same metric is used in the W-algebra for fuzzy regression [20]. Since the fuzzy computations reduce to three real-valued recurrences with shared coefficients, all operations are differentiable, and training is performed with an ordinary gradient method, in our case Adam [23]. For the same reason, the associative parallel scan [24] on which fast Mamba training is based applies to recurrence (6) without changes. The computational cost grows threefold compared with the crisp model, while the asymptotic complexity $O(TDN)$, linear in the history length, is preserved. In real-time operation the customer's state is updated after each week in $O(DN)$ operations, and the purchase history does not need to be stored.

It remains to translate the fuzzy forecast into language understandable to the decision maker. We define on $[0,1]$ the linguistic terms of low, medium and high preference by triangular membership functions $\mu_{T_k}$. The degree to which a forecast matches a term is naturally assessed by the possibility measure

$$\Pi(\tilde{p} \text{ is } T_k) = \sup_z \min\Big(\mu_{\tilde{p}}(z), \, \mu_{T_k}(z)\Big), \tag{11}$$

which shows how compatible the forecast is with the given description. The change of these quantities over time yields statements about preference dynamics, for example that the preference is shifting from high to medium, and thus points to customers who need attention.

## 4. Computational experiment

To test the model, we consider a synthetic problem in which the true preference is known and we can assess how well the model recovers it. We simulate 2500 customers over 52 weeks and the preference for one brand within a product category. The true preference $\theta_t$ relaxes towards an individual long-run level $\theta^*$ distributed as $\mathrm{Beta}(6,2)$ according to $\theta_{t+1} = \theta_t + 0.25(\theta^* - \theta_t) + \varepsilon_t$ with Gaussian noise $\varepsilon_t$ of standard deviation 0.02 (values are restricted to $[0.01, 0.99]$). In randomly chosen weeks, with probability 0.15 per week, a competitor runs a promotion announced one week in advance. During a promotion the customer's preference temporarily drops to $\theta_t(1 - 0.6\sigma_i)$, where $\sigma_i$ is the customer's promotion sensitivity, uniformly distributed on $[0,1]$. In addition, with probability $0.12\sigma_i$ a promotion causes an irreversible switch of the customer, in which the long-run level is replaced by a value drawn from $\mathrm{Beta}(2,6)$. The switch takes effect in the same week, but the preference does not drop abruptly; it relaxes to the new level with a characteristic time of about four weeks. The number of purchases in the category per week is Poisson with mean $\lambda_i(1 + 0.3\sin(2\pi t/26))$, where the intensity $\lambda_i$ is gamma-distributed with mean 3, and the number of purchases of the brand is binomial with probability equal to the current preference.

Weekly observations are converted into triangular numbers by a simple rule that mimics an expert's reasoning that the brand's share is about $p$, but the fewer the purchases, the less certain the estimate. With $k$ purchases of the brand out of $n$ purchases in the category, we set $p = (k+1)/(n+2)$, $s = 2\sqrt{p(1-p)/(n+2)}$ and $\tilde{x}_t = \Pi_{[0,1]}(p - s,\, p,\, p + s)$. With no purchases this gives the wide number $(0,\ 0.5,\ 1)$, which expresses the absence of definite information.

The developed selective model FW-Mamba was compared with four alternatives. The naive forecast repeats the last observation. The evolution operator with constant inertia $\tilde{h}_t = \lambda\tilde{h}_{t-1} + (1-\lambda)\tilde{x}_t$ is a special case of (6) without selectivity; the learned inertia was $\lambda \approx 0.82$. To separate the contribution of external factors from that of the selective architecture, we also trained FW-Mamba without external factors, in which the promotion and seasonality features are set to zero. Finally, to assess the cost of fuzziness, we trained a crisp model of the same architecture on the observation centres $c(\tilde{x}_t)$ without the width feature. All models were trained on 2000 customers and tested on 500 other customers. FW-Mamba variants were trained for 40 epochs, and the constant-inertia operator, which has a single parameter, for 20 epochs. The whole experiment was repeated on three independent data generations.

Quality was assessed by several measures. The accuracy of the forecast centre is characterised by the mean absolute error of the centre $c(\tilde{p}_{t+1})$ with respect to the next week's true preference, which is known in the synthetic problem. Closeness to the fuzzy observation is characterised by the root-mean-square distance (10). The consistency of the fuzzy forecast with the truth is characterised by the mean membership degree $\bar{\mu}$ of the true value in the forecast, by the fraction of cases in which the true value falls within the support $[l, r]$, and by the mean support width. The results are given in Table 1.

Table 1. Results on test customers, mean and standard deviation over three data generations

| Model | Centre error | Distance (10) | $\bar{\mu}$ | Coverage | Width |
|---|---|---|---|---|---|
| Naive forecast | 0.179 ± 0.002 | 0.198 ± 0.003 | 0.546 | 0.954 | 0.753 |
| Constant inertia | 0.163 ± 0.003 | 0.152 ± 0.001 | 0.568 | 0.988 | 0.750 |
| FW-Mamba without external factors | 0.163 ± 0.003 | 0.151 ± 0.001 | 0.552 | 0.974 | 0.731 |
| Crisp model of the same architecture | 0.155 ± 0.003 | — | — | — | — |
| FW-Mamba | 0.154 ± 0.003 | 0.147 ± 0.001 | 0.578 | 0.986 | 0.729 |

The developed FW-Mamba model gives the smallest centre error and the smallest distance to the fuzzy observation. Comparison with the row without external factors shows where the gain comes from. Without information about promotions, the selective model merely reproduces the result of the constant-inertia operator, and the reduction of the error from 0.163 to 0.154 is entirely due to the use of external factors. Fuzziness does not degrade the accuracy of the centre. The crisp model of the same architecture gives an error of 0.155, which coincides with the result of the fuzzy model within the spread. At the same time, the fuzzy model additionally reports how reliable the forecast is. The true value falls within the forecast support in 98.6% of cases, and the mean membership degree of the true value is higher than for all other fuzzy variants, with the smallest support width. As follows from Section 3, all obtained forecasts are proper triangular numbers in $[0,1]$.

The small number of parameters makes it possible to understand how the model uses information about promotions. The learned crisp shift $\beta_t$ before an announced competitor's promotion equals $-0.08$ on average, whereas in ordinary weeks it is close to $+0.01$. This means that the model lowers its preference forecast for the promotion week in advance. The discretisation step $\Delta_t$ before a promotion is larger than in ordinary weeks in two data generations out of three (0.32 versus 0.26 and 0.28 versus 0.24), that is, the model updates its state faster, but in the third generation there is no difference (0.28 versus 0.29). Consequently, in this example the influence of external factors is transmitted mainly through the shift, while selective memory plays a secondary role. This agrees with the structure of the data, in which a promotion primarily causes a short-term dip in preference.

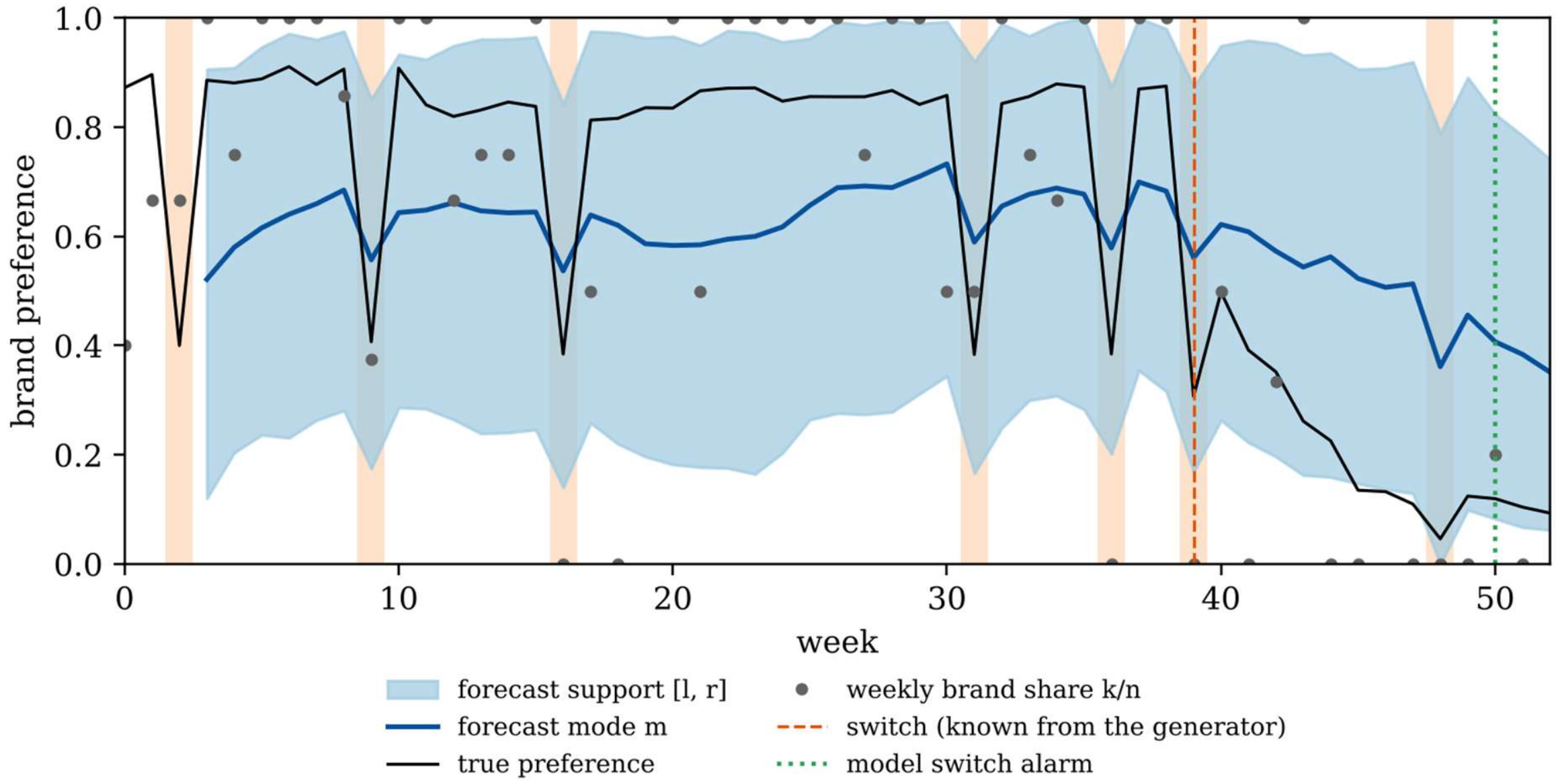

Fig. 1. One-week-ahead FW-Mamba forecast for a test customer for whom a competitor's promotion in week 39 caused a permanent switch. The switch week is known from the data generator and is marked for comparison; it is not given to the model. Orange marks the weeks of competitor promotions, the band shows the support of the triangular forecast, the blue line shows its mode, and the green dotted line marks the week in which rule (12) raises a switch alarm.

Figure 1 shows the model's operation for a customer for whom a competitor's promotion in week 39 caused an irreversible switch. The model reproduces the temporary preference dips during promotions and gradually lowers the forecast after the switch. The linguistic interpretation by formula (11) with the terms low $(0, 0, 0.45)$, medium $(0.2, 0.5, 0.8)$ and high $(0.55, 1, 1)$ preference is given in Table 2. The possibility of high preference decreases from 0.60 to 0.38, while the possibility of low preference increases from 0.12 to 0.43. For a manager this means that the customer's preference is shifting from medium tending to high to medium tending to low, and the customer is entering the risk zone.

Table 2. Linguistic interpretation of the forecast for the customer shown in Fig. 1

| Week | Forecast (l; m; r) | Π(low) | Π(medium) | Π(high) | Assessment |
|---|---|---|---|---|---|
| 37 | (0.35; 0.70; 1.00) | 0.12 | 0.69 | 0.60 | medium, closer to high |
| 43 | (0.16; 0.54; 0.93) | 0.34 | 0.93 | 0.46 | medium, decreasing |
| 49 | (0.10; 0.46; 0.89) | 0.43 | 0.94 | 0.38 | medium, closer to low |

It should be emphasised that the switch week in Fig. 1 is taken from the data generator rather than determined by the model, and at the moment of the switch it cannot be established either from the observations or from the forecast. The switch is caused by a promotion, and in week 39 the true preference dips in the same way as during previous promotions. The difference becomes visible only in the following weeks. After earlier promotions the preference returned to its previous level of about 0.85 within one week, whereas after week 39 it recovers only partially and then keeps decreasing towards the new level. From week 41 on, zeros prevail among the observed purchase shares. Therefore a temporary dip can be distinguished from a permanent shift only by several subsequent observations, and any way of detecting a switch, whether visual inspection of the plot or a model, inevitably works with a delay.

The linguistic interpretation of the forecast makes it possible to formulate a simple switch detection rule that requires no additional training. We assign the forecast for week $t$ to the state “rather low” if the possibility $\Pi(\tilde{p}_t \text{ is low})$ is not less than the possibility $\Pi(\tilde{p}_t \text{ is high})$, and to the state “rather high” otherwise. Weeks for which a competitor’s promotion has been announced are skipped, since the forecast decrease in them is expected and temporary. A switch alarm is raised in week $t^{\mathrm{a}}$ if the customer was previously in the state “rather high” and in the last two weeks without promotions, including week $t^{\mathrm{a}}$, the condition

$$\Pi(\tilde{p}_t \text{ is low}) \geq \Pi(\tilde{p}_t \text{ is high}) \quad (12)$$

holds. The rule was applied to test customers who switched between weeks 4 and 40 and were in the state “rather high” before the switch. An alarm for a customer without a switch who had been in the state “rather high” at least once was counted as a false alarm. The results for FW-Mamba and the constant-inertia operator are given in Table 3.

Table 3. Switch detection by rule (12) on test customers. Fractions are averaged over three data generations; for the delay, the range of medians over generations is given

| Model | Detected | Median delay, weeks | Early alarm | False alarms |
|---|---|---|---|---|
| Constant inertia | 87% | 7–8 | 9% | 31% |
| FW-Mamba | 88% | 7–9 | 4% | 22% |

Both models detect most switches, FW-Mamba in 88% of cases. The median alarm delay is 7–9 weeks for FW-Mamba and 7–8 weeks for the constant-inertia operator, with the interquartile range from 6 to 11 weeks in both cases. This delay has two causes. The first is fundamental: after a switch, the true preference relaxes to the new level with a characteristic time of about four weeks, and with three purchases per week a permanent shift can be distinguished from random fluctuations only by several observations. The second is related to the model’s inertia and to the smoothing in the fuzzification rule, because of which the forecast mode has to fall to about 0.5 before the possibility of low preference exceeds that of high preference. For the customer in Fig. 1 the alarm is raised in week 50. The main difference between the models lies in the number of erroneous alarms. Taking announced promotions into account reduces the fraction of early alarms raised before the actual switch from 9% to 4%, and the fraction of false alarms for customers without a switch from 31% to 22%. The constant-inertia operator does not know about the upcoming promotion and takes the dip it causes for the beginning of customer churn, whereas FW-Mamba accounts for the promotion through the shift $\beta_t$. The remaining false alarms occur mainly for hesitant customers whose long-run preference is close to the boundary between the low and high preference terms. Rule (12) is deliberately simple and serves to illustrate that the linguistic interpretation of the forecast is suitable for managerial decisions. It can be refined, for example, by comparing the current forecast with the customer’s own past level or by using the model’s long-memory channels as a reference level.

The example also shows the limitations of the approach in its current form. The forecast mode for loyal customers is systematically closer to 0.5 than the true preference. The reason is the smoothing $p = (k + 1)/(n + 2)$ in the fuzzification rule, since the model is trained to reproduce the fuzzy observation rather than the hidden true value. In addition, the fuzzification rule converts a statistical estimate of the share and its standard error into a triangular number, that is, it mixes the frequentist and possibilistic interpretations of uncertainty. Such mixing is acceptable as an engineering heuristic, but the support $[l, r]$ is then not a confidence interval, and the coverage of 98.6% in Table 1 characterises the consistency of the forecast with the truth in this example rather than a guaranteed probability. Finally, rule (12) depends on the choice of the boundaries of the linguistic terms, and its sensitivity to them requires separate analysis.

## 5. Conclusion

We have proposed the fuzzy selective state space model FW-Mamba for forecasting the dynamics of consumer preferences. A preference is described by a triangular fuzzy number, and its evolution is described by a lightweight Mamba-type selective recurrence computed in Matveev's two-component algebra. The central element of the construction is the separation of roles, in which the crisp part of the model decides how much to update the representation of the customer, while the fuzzy part stores this representation together with a measure of its vagueness. Owing to this, the fuzzy computational path contains only non-negative linear operations, shift and clipping. For such operations Matveev's algebra coincides with Zadeh's extension principle, the model state remains a proper triangular number and is a convex combination of past observations, and the fuzziness of the forecast does not accumulate over time. The computational advantages of Mamba are retained, including training complexity linear in the history length and constant-time updates of the customer state.

The choice of a state space model instead of a transformer is motivated not only by computational efficiency. The recurrent core of Mamba is linear and admits exact propagation of fuzziness, whereas in the attention mechanism the fuzziness would have to be approximated.

The synthetic example shows that the 107-parameter model outperforms the constant-inertia evolution operator and the naive forecast, is as accurate as a crisp model of the same architecture and provides an interpretable linguistic assessment of the forecast. The linguistic interpretation of the forecast makes it possible to detect a permanent switch of a customer to a competitor with a simple rule, and taking announced promotions into account reduces the number of false alarms by about one third compared with the constant-inertia operator. The detection delay of several weeks is explained by the fact that a switch can be distinguished from a temporary dip only by subsequent observations. At the same time, the example is one-dimensional, the data are synthetic, and the fuzzification rule is simple and symmetric. Further work includes validation on open retail transaction datasets, extension to multidimensional preferences over several categories and brands, the use of asymmetric fuzzy observations, the study of fuzzy weights admitted by Matveev's algebra, and comparison with Mamba4Rec, SIGMA, transformers and probabilistic methods of uncertainty estimation.